\documentclass[12pt,a4paper]{article}

\usepackage{graphicx}% Include figure files
\usepackage{dcolumn}% Align table columns on decimal point
\usepackage{bm}% bold math
\usepackage{color}
\usepackage{epsf}

\begin{document}

\title{Long-term neuronal behavior caused by two synaptic modification mechanisms}

\author{Xi Shen (1), Philippe De Wilde (2)
\\((1) Department of Electrical and Electronic Engineering,\\
Imperial College London, United Kingdom, \\(2) Department of
Computer Science, \\Heriot-Watt University, United
 Kingdom)}

\date{\today}

\begin{abstract}

We report the first results of simulating the coupling of
neuronal, astrocyte, and cerebrovascular activity. It is suggested
that the dynamics of the system is different from systems that
only include neurons. In the neuron-vascular coupling,
distribution of synapse strengths affects neuronal behavior and
thus balance of the blood flow; oscillations are induced in the
neuron-to-astrocyte coupling.

\end{abstract}

\maketitle

Recently, more details have emerged of the interaction between
neurons, glial cells, and the cerebrovascular network
\cite{ph:one, ph:two, ph:three}. Most of this work is on the
micro-level, involving only a few cells and capillaries or
arterioles. Numerous investigations have confirmed the following
discoveries: The modulation of synaptic efficacy will affect
emergent behavior of neuronal assemblies \cite{ph:ten, ph:eleven};
Dilation of capillaries is highly related to the activity of
nearby neurons \cite{ph:thirteen}; Astrocytes are very sensitive
to the level of neuronal activity because of their position and
their sensitivity to activity-dependent changes in the chemical
environment they share with neurons \cite{ph:nine}; Intercellular
Calcium waves between astrocytes are the main signalling mechanism
within glial cell networks \cite{ph:six, ph:seven}. In this
letter, we will build numerical models based on those
physiological findings, because micro-descriptions using
simplified models of firing \cite{ph:twelve, prl:two} and wave
propagation can be inserted into larger scale simulations. Here we
show a modification of the synapse strengths that allows the
neuronal firing and the cerebrovascular flow to be compatible on a
meso-scale; with astrocyte signalling added, limit cycles exist in
the coupled networks.

Neurons are associated with capillaries in the brain, and
according to physiological discoveries, the neuronal activity can
dilate the capillaries that supply them. Our first model contains
2400 neurons and 30 branches of capillaries, each of which
supplies 80 neurons. Each neuron has two states: '1' indicates
that at that time the neuron was firing and '-1' means the neuron
was not firing. All the synapses between these neurons are
described by a 2400 by 2400 matrix $S$, in which each component
$S_{i,j}$ represents the strength of the synapse from neuron $j$
to neuron $i$. This definition is according to the Hopfield
artificial neural network model \cite{ph:fourteen}. On average,
each neuron has a certain number of synapses which are linking to
other neurons. We suppose that the synapse number follows a normal
distribution that peaks at $S_{ave}$. As a result, most of the
neurons have a similar number of synapses. This is based on the
assumption that a small area in the brain is uniform.

These 2400 neurons can be viewed as a very small part of the
brain, in which each neuron has synapses connected with other
neurons within this part as well as outside of this part. Here we
define another parameter $p_{loc}$, which indicates the locality
of the synapses, i.e., how many of one neuron's synapses are
connected to the neurons within this area. Consequently, the input
signal of each neuron consists of two parts: one part is from the
local neurons which are connected with it via synapses and the
other part is from the neurons outside, this part can be regarded
as Gaussian noise. The capillary model is simple, the original
blood flow is set as follows: the blood flow of the first layer
(top two branches) is 800 $\mu m^{3}/ms$ , and where the branches
split, both the directions have half this flow, $400 \mu
m^{3}/ms$; this rule is applied all the way to the bottom layer,
where all the 16 branches have a flow of $100 \mu m^{3}/ms$, which
is the low velocity of flow in the capillaries quoted from data of
blood circulation \cite{ph:fifteen}. If the number of firing
neurons associated with this capillary exceeds a certain threshold
at time $T$, the capillary will dilate at the time, which means in
our model that the blood volume is four times as large as the
original one. In addition, the dilation period is set to $d_{c}$,
because the change of capillary width is slower than that of the
neuron potentials. We also define a parameter which represents the
compatibility as follows:

\begin{equation}
Cmp=\sum_{j=1}^{j=14}|Bldflw_{j}-Bldflw_{2j+1}-Bldflw_{2j+2}|
\end{equation}

Compatibility is the summation of the blood flow difference over
all 14 capillary joint nodes; the capillary network is illustrated
in Fig. 1. We call a network compatible if the compatibility
(Eq.(1)) is zero.

\begin{figure}
\begin{center}
\includegraphics[width=3.2in]{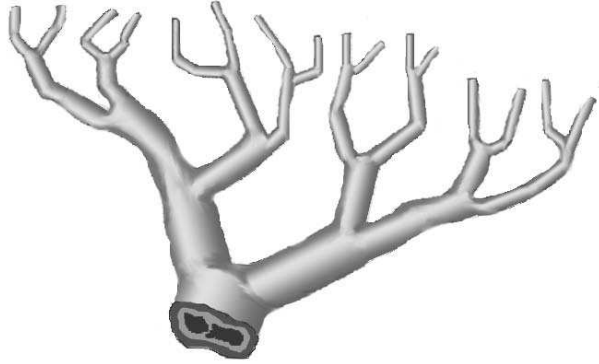}\\
\caption{Thirty branches of capillaries.}
%longer caption
%add label
\end{center}
\end{figure}

\begin{figure}
\begin{center}
\includegraphics[width=3.2in]{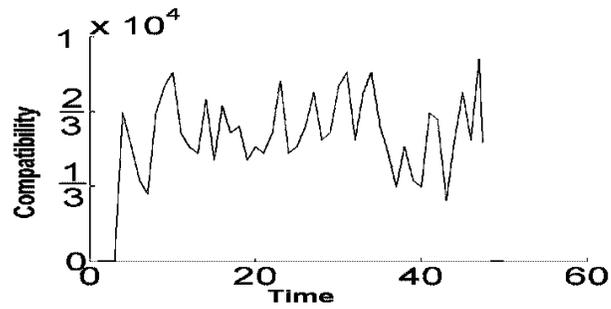}\\
\caption{Compatibility for the model with synaptic weights
uniformly distributed between -1 and 1. The blood flow is not
compatible.} \label{fig:two}
\end{center}
\end{figure}

The dilation of each capillary is caused by the fluctuation of the
firing number associated with it, i.e., if more than $N_{firing}$
neurons that this capillary supplies are firing, this capillary is
going to be dilated after $\Delta t$ time steps and this dilation
will last $d_{c}$ time steps. In a first set of simulations, all
the synapse strengths are uniformly distributed between -1 and 1,
the blood flow turned out to be highly incompatible, since the
blood flow difference fluctuated between 3000 and 8000 $\mu
m^{3}/ms$ (also see Fig. ~\ref{fig:two}).

In the second simulation, we manually set these strengths to be
uniformly distributed between $2\mu-1$ and 1, i.e., the mean value
of them is $\mu$, which we call local-shift. Initially, the neuron
states are set according to the rule that 1/4 of the neurons are
firing. The simulation results are shown in Fig. 3. We found that
this initialization had no effect on the behavior of the system;
neurons will reach their resting state, which is determined by
$\mu$. Over a certain short time period (from T=101 to T=105 in
this simulation), we increased the Gaussian noise in this area so
this positive input forces a number of neurons to fire. Then the
neurons' state will be changed to 'excitatory', and at the same
time the high level of neuron firing will trigger the dilation of
capillaries. The results imply that this temporary external noise
triggers the 'excitatory' state and that a local shift $\mu$ helps
the maintenance of this state until the next inhibitory signal
arrives. This quick change from 'resting' to 'excitatory' makes
the incompatible time really short, so in this way, these coupled
networks function well.

\begin{figure}
\begin{center}
\includegraphics[width=3.2in]{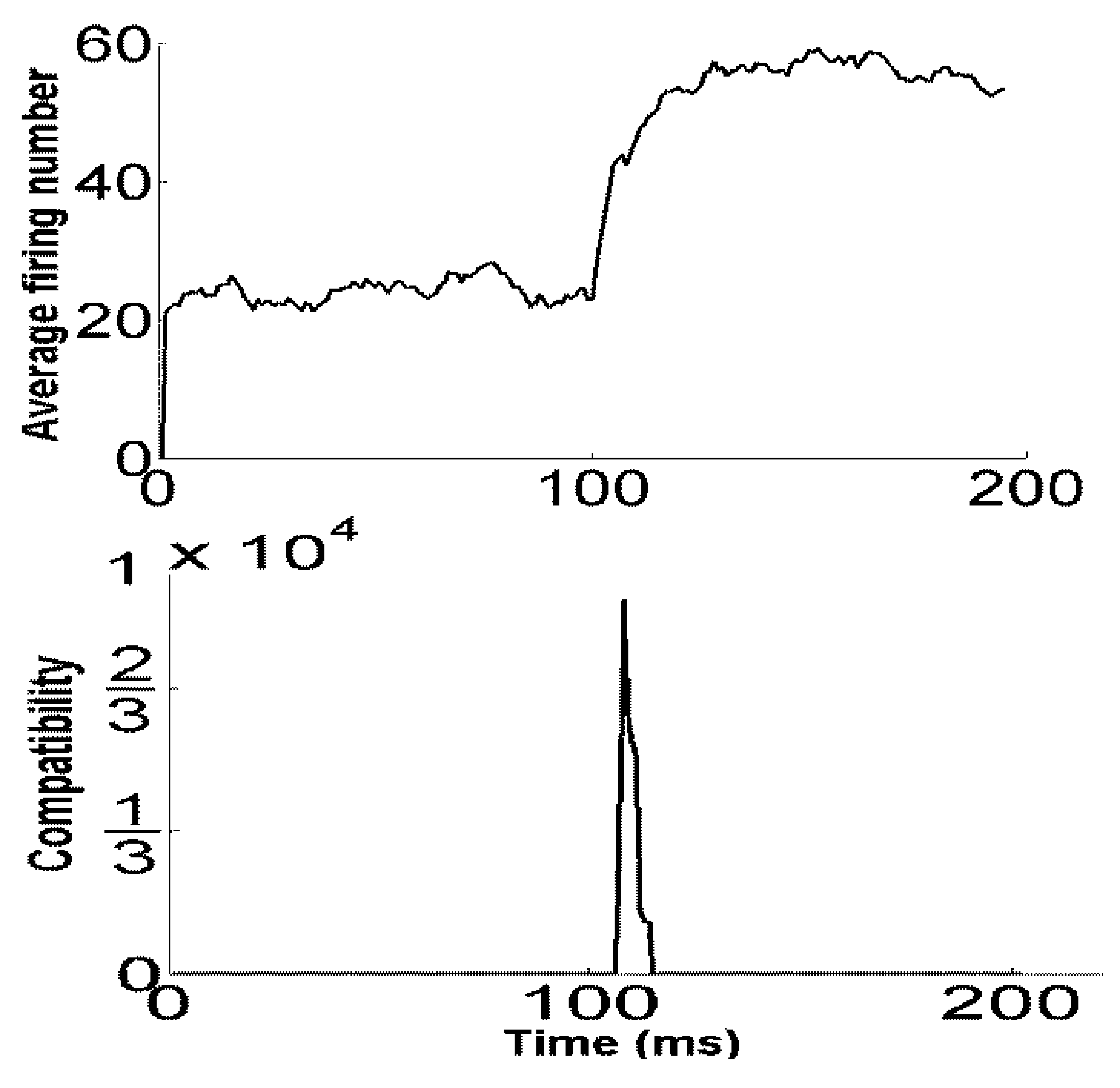}\\
\caption{Average firing number per branch (top): We calculated the
firing number associated with each branch of capillary, evaluated
the mean value and plotted in this figure. Compatibility (bottom):
The incompatible time is really short because the neurons quickly
change from 'resting' to 'excitatory'.}
\end{center}
\end{figure}

Based on these results, we can make a few hypotheses about the
neurovascular coupling in the brain. The first hypothesis is that
the local synapses are likely to have positive strengths. This
enables the cluster of neurons to maintain either the 'excitatory'
state or the 'resting' state, and external signals cause the
switch between these two states. From simulations we can see that
this switch is fast, although the capillaries react with time
delay, the incompatible time is so short that it can be ignored.
Secondly, since the local shift of the synapse weights is so tiny,
we could assume that this is associated with the neuronal
communication. The communication between neurons could slightly
change the strength of synapses connected them, and this change is
on the same time scale as the dilation of the capillaries. Because
the neuron state is determined by this local-shift, changing of
strength could result in changing of state, and at the same time
balance the blood flow.

It is now believed that glial cells (astrocytes) also have their
own network (via Calcium waves) in the brain and that this network
also plays a distinct role in information processing. Calcium
waves can both propagate among networks of astrocytes globally and
send specific Calcium signals to a small number of nearby
astrocytes \cite{ph:sixteen}. In other words, glial signalling
also has its preferred routes. Communication between glial cells
and neurons is bidirectional and complex. Astrocytes are very
sensitive to the level of neuronal activity because of their
position and their sensitivity to activity-dependent changes in
the chemical environment shared by neurons and astrocytes
\cite{ph:nine, ph:eight}. Astrocytic Calcium waves can be
triggered by synaptically-released neurotransmitters and in
addition the frequency of these Calcium oscillations can change
according to the level of synaptic activity \cite{ph:seventeen}.
It is observed that activated astrocytes can control the synaptic
transmission by regulating the release of neurotransmitter from
the nerve terminal. This regulation can be either excitatory (by
secreting the same neurotransmitter) or inhibitory (by absorbing
the neurotransmitter). If regulation of synaptic activities is the
short-term effect of astrocytes, they can also change long-term
synaptic strengths by releasing signalling molecules that cause
the axon to increase or decrease the amount of neurotransmitter
\cite{ph:sixteen}. To sum up briefly, it is now possible to model
the neuron-to-astrocyte coupling because many recent experiments
confirm the signalling pathways among these two types of networks.
Neurons communicate with neurons while astrocytes 'listen to' the
neuronal communication. Basing upon what they 'hear', astrocytes
regulate neuronal activities and communicate with other
astrocytes. In order to quantify these regulations, we will define
rules dependent on several parameters and then simulate this
model.

The neuronal network contains 900 neurons, each of which has
either a '1' state (firing) or a '0' state (resting). There are
3600 astrocytes in the glial network. In order to simplify, we
suppose these astrocytes are in a $60 \times 60$ matrix $G$. The
state of an astrocyte is $g_{i,j}$ and it could be zero or a
positive number. These 3600 astrocytes have their own connections,
too. Because Calcium signals are more likely to propagate among
nearby astrocytes, we assume here that the connection between
astrocyte $(i1, j1)$ and $(i2, j2)$ is given by probability
$\rho$, which is inversely proportional to the distance between
them ($\sqrt{(i1-i2)^{2}+(j1-j2)^{2}}$). Since these connections
are in fact invisible Calcium signal pathways, we did not give
them weights. Unlike electrical signals travelling between
neurons, Calcium signals need more time to travel from one
astrocyte to another, and we set this time delay proportional to
the distance.

The coupling between the neuronal network and the glial network is
complicated. We will assume that each neuron has four astrocytes
nearby which are all sensitive to this neuron's activity. At the
same time, because in the brain a single astrocyte can cover a
large number of synapses \cite{ph:seven}, in our model we allocate
to each astrocyte 1/4 of all the afferent synapses of this neuron,
so that four nearby astrocytes can cover all afferent synapses
without overlapping. Astrocytes can sense the spike frequency
along each synapse, regulate the release of neurotransmitters, and
change the synaptic strength. A simplified illustration of the
position of neurons and astrocytes can be found in Fig. 4. There
are two factors which affect the state of an astrocyte. The first
one is high frequency effective spikes along synapses that this
astrocyte is sensing, and 'effective' means either a positive
spike making the target neuron fire or a negative spike making it
rest. The second factor is the Calcium signals from other
astrocytes, and this part can accumulate as well. If the input
signal of an astrocyte exceeds a certain threshold , this
astrocyte will become active, which means that the state of the
astrocyte is nonzero. Since activated astrocytes can have
different levels of activation, we define that when activated, an
astrocyte's state equals its input.

\begin{figure}
\begin{center}
\includegraphics[width=2.6in]{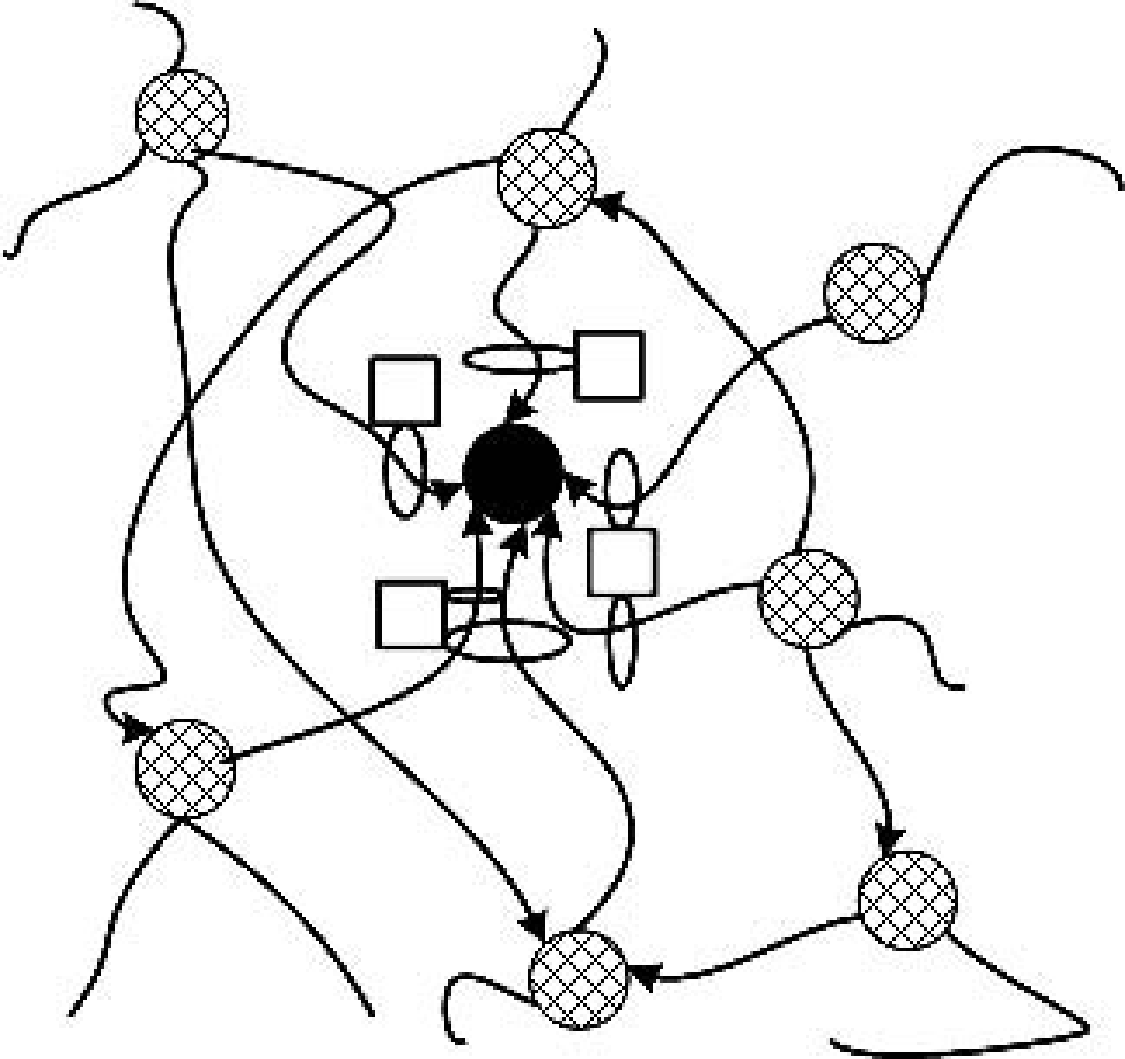}\\
\caption{The position of one neuron (black circle) and 4 nearby
astrocytes (rectangles). To make it clear, only afferent synapses
from other neurons (shadowed circle) are drawn. The arms of
astrocytes (ellipses) surround some synapses to sense and regulate
their activities.}
\end{center}
\end{figure}

As mentioned previously, activated astrocytes can regulate the
release of neurotransmitters over short periods of time ($<50s$)
\cite{ph:eighteen}, in other words; they can change the input
signal of the nearby neuron within a few seconds. Because the
conditions for excitation and inhibition are unknown, we will make
the following assumptions. The neuron's input will be increased by
an activated astrocyte proportionally to its state, but if this
astrocyte was over-activated (its state exceeded a certain value),
it would inhibit the input of the neuron. In the brain this local
rule may be more complicated, and we believe that this interaction
between neurons and astrocytes is crucial to short-term memory and
information processing.

Meanwhile, as a long-term effect, astrocytes can also monitor the
synaptic activities and gradually modify the strengths of synaptic
connections (efficacies). This mechanism is also called adapting
synapses, and has been studied before \cite{ph:nineteen}. Because
our model is based on discrete time, the update rule of synaptic
strengths is quite simple. For a positive synapse, we look back 20
time steps and calculate the spike frequency of this synapse and
the firing frequency of the neuron. If both of the two frequencies
are high and correlated, in other words, the high firing rate is
triggered by episodes of high synaptic activity through this
connection (we call it excitation success), we will make this
connection stronger. On the other hand, if this synapse is
negative, it will be enhanced if the spike frequency is high but
the neuron is firing at a low rate (inhibition success). The
strength can also be weakened if high spike frequency along a
positive synapse does not cause high firing rate (excitation
failure) or high spike frequency along a negative synapse does not
cause low firing rate (inhibition failure). Although due to
limitations of computing time, our simulating time can not be long
enough to represent the gradual changing of efficacy; we still
include this procedure in our program. We believe that the
distribution of synaptic strengths heavily affects the emergent
behavior of neuron-to-astrocyte coupling, so it is crucial in
understanding the brain function, especially long-term memory and
learning \cite{ph:twenty,ph:twentyone}.

\begin{eqnarray}
n_{p}(t+1)=\frac{1}{2}sgn(\sum_{q=1}^{900}T_{qp}(t)n_{q}(t)+\sum_{r=1}^{4}g_{4p+r}(t)-\varphi_{p}(t)+\eta)+\frac{1}{2}
\end{eqnarray}

Finally, we will explain the time resolution of this model. Based
on the absolute refractory period of neurons, we suppose that each
time step in our simulation equals 1 ms in the real brain. Since
neurons can not be firing at two consecutive time steps, the
maximum firing frequency is 500 Hz. At each time step, we loop
over all astrocytes, all neurons and all synapses and change their
states according to the rules we have defined above. The update
rule for a neuron can be found in Eq.(2), in which $T_{qp}(t)$
represents the time-dependent strength,
$\sum_{r=1}^{4}g_{4p+r}(t)$ means contributions from astrocytes
and $\varphi_{p}(t)$ and $\eta$ are the threshold and noise,
respectively.

\begin{figure}
\begin{center}
\includegraphics[width=3.2in]{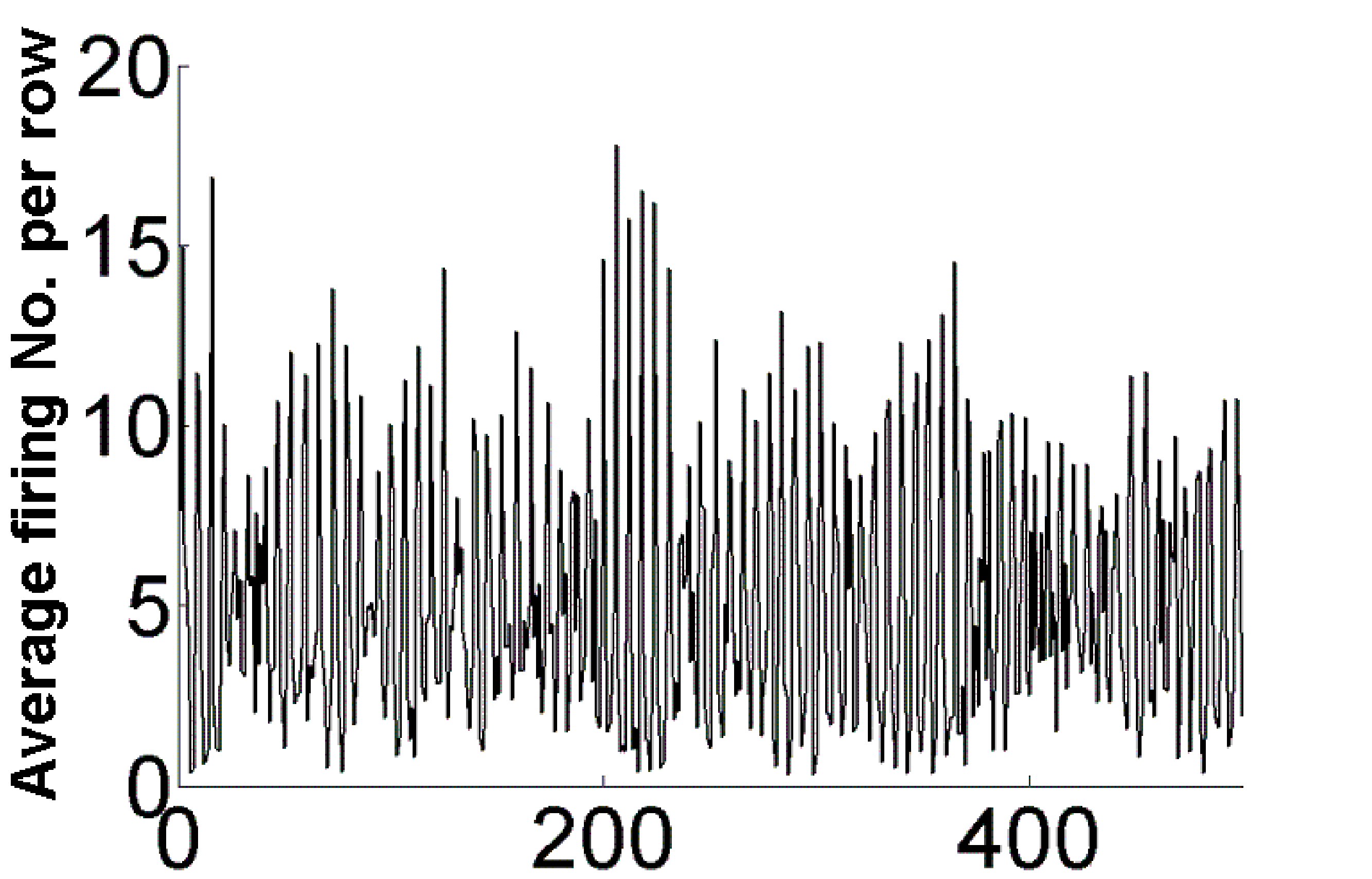}\\
\includegraphics[width=3.2in]{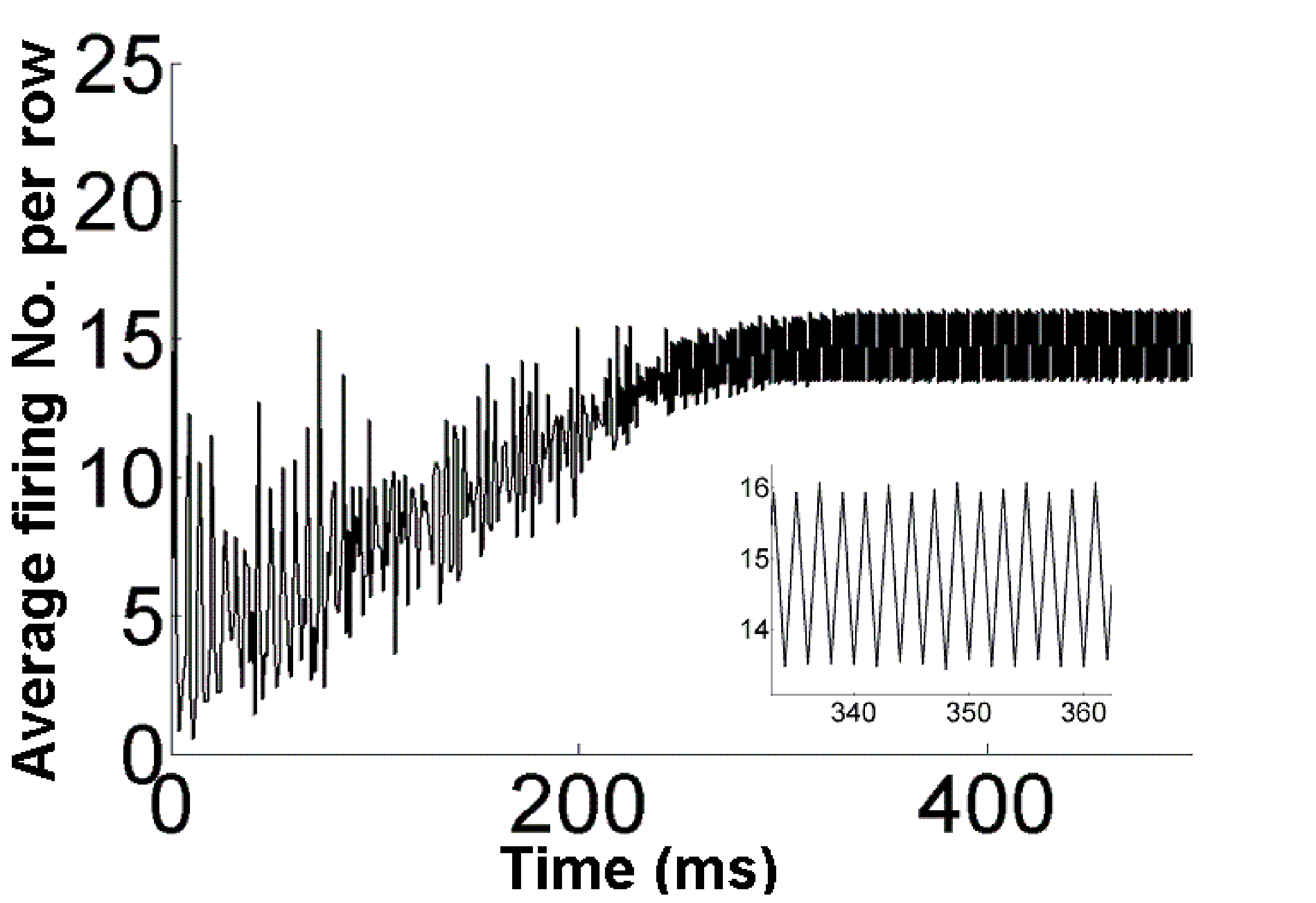}\\
\caption{Top figure: Average number of firing neurons in each row
in our simulation of pure neurons. Bottom figure: average number
of firing neurons in each row in our simulation of
neuron-to-astrocyte coupling.}
\end{center}
\end{figure}

Our results (Fig.5) imply that in the pure neuronal network,
neuronal activity was quite random within the whole 500 time
steps. While in the simulation with both neurons and astrocytes,
it is obvious that there was an attractor in the neuronal network,
after T=300, the neuronal network just oscillated between two
states. If we compare the two figures concerning average firing
number in Fig. 5, it is easy to find out that the firing frequency
of neurons in the coupled network is much higher than that in the
pure neuronal network. According to our simulations, on average 17
percent of all neurons were firing in the pure neuronal network
while in the coupled network it was 39.2 percent. This is mainly
because astrocytes changed the strengths of many synapses. At the
same time, activated astrocytes enhancing the afferent spikes play
a small part in this as well.

We have shown that a modification of the synapse strengths can
allow the neuronal firing and the cerebrovascular flow to be
compatible on a meso-scale. With astrocyte signalling added, limit
cycles exist in the coupled networks. This is a first step towards
a better understanding of the coupling of neuronal, astrocyte and
cerebrovascular networks in the brain. This coupling is essential
in any modern description of the brain, and has to be investigated
using methods of complex systems. They will ultimately show the
right scale for the models, and whether critical phenomena exist.

\end{document}